\documentclass[10pt, conference]{IEEEtran} 
\IEEEoverridecommandlockouts

\usepackage{cite}
\usepackage{amsmath,amssymb,amsfonts}
\usepackage{algorithmic}
\usepackage{graphicx}
\usepackage{textcomp}
\usepackage{xcolor}
\usepackage{booktabs}
\usepackage{pgfplots}

\usepackage[colorinlistoftodos,prependcaption,textsize=small]{todonotes}
\usepackage{regexpatch}
\makeatletter
\xpatchcmd{\@todo}{\setkeys{todonotes}{#1}}{\setkeys{todonotes}{inline,#1}}{}{}

\usepackage{url}
\usepackage{svg}

\usepackage{cleveref}
\usepackage{flushend}

\def\BibTeX{{\rm B\kern-.05em{\sc i\kern-.025em b}\kern-.08em
    T\kern-.1667em\lower.7ex\hbox{E}\kern-.125emX}}

\newcommand{\ic}[1]{\raisebox{-.5\height}{#1}}
    
\DeclareMathAlphabet{\mathcal}{OMS}{cmsy}{m}{n}
\graphicspath{{images/}}

\makeatletter
\def\ps@IEEEtitlepagestyle{%
  \def\@oddfoot{\mycopyrightnotice}%
  \def\@evenfoot{}%
}
\def\mycopyrightnotice{%
  \begin{minipage}{\textwidth}
  \centering \scriptsize
  Copyright~\copyright~2023 IEEE. Personal use of this material is permitted. Permission from IEEE must be obtained for all other uses, in any current or future media, including\\
  reprinting/republishing this material for advertising or promotional purposes, creating new collective works, for resale or redistribution to servers or lists, or reuse of any copyrighted component of this work in other works.
  \end{minipage}
}
\makeatother

\begin{document}

\title{\LARGE \bf Utilizing Hybrid Trajectory Prediction Models to Recognize Highly Interactive Traffic Scenarios}

%%%%%%
\author{
Maximilian~Zipfl$^{1,2,\dagger}$,
Sven~Spickermann$^{2, \dagger}$, 
and J.~Marius~Zöllner$^{1,2}$
\thanks{$^{1}$FZI Research Center for Information Technology, Karlsruhe, Germany
{\tt\small zipfl@fzi.de}}%
\thanks{$^{2}$Karlsruhe Institute of Technology, Karlsruhe, Germany}%
\thanks{$^{\dagger}$  Both authors contributed equally to this work as first authors.}
}%

\maketitle

\begin{abstract}
Autonomous vehicles hold great promise in improving the future of transportation. The driving models used in these vehicles are based on neural networks, which can be difficult to validate. However, ensuring the safety of these models is crucial. Traditional field tests can be costly, time-consuming, and dangerous. To address these issues, scenario-based closed-loop simulations can simulate many hours of vehicle operation in a shorter amount of time and allow for specific investigation of important situations. Nonetheless, the detection of relevant traffic scenarios that also offer substantial testing benefits remains a significant challenge. To address this need, in this paper we build an imitation learning based trajectory prediction for traffic participants. We combine an image-based (CNN) approach to represent spatial environmental factors and a graph-based (GNN) approach to specifically represent relations between traffic participants. In our understanding, traffic scenes that are highly interactive due to the network's significant utilization of the social component are more pertinent for a validation process. Therefore, we propose to use the activity of such sub networks as a measure of interactivity of a traffic scene. We evaluate our model using a motion dataset and discuss the value of the relationship information with respect to different traffic situations.

\end{abstract}

% ==============
%  Introduction
% ==============
\section{Introduction}
The development of fully autonomous vehicles promises potential solutions to a wide range of problems of an ecological, economic and social nature. However, for such goals to be achieved at all, the safety and reliability of the driving models currently under development must be ensured.
Scenario-based testing holds great potential due to the continuous improvement of computing power and simulation environments.
Identifying relevant and meaningful traffic scenarios for the scenario-based testing procedure is a critical research question since it is not feasible to test all possible scenarios due to their infinite nature. Hence, test engineers need to limit their selection to scenarios that are likely to be significant and informative for the validation process.

Several approaches have been devised in recent years to aid in defining and reducing the so-called test space. These methods encompass a broad range of techniques, including parameter exploration and scenario evaluation using criticality metrics \cite{zipfl_fingerprint_2022, sun_scenario-based_2022, westhofen_criticality_2023}.

Our hypothesis is, that traffic scenarios that involve a high degree of social interaction, where careful attention to other traffic participants is required to establish the appropriate trajectory, hold a significant value for scenario-based testing.
In the context of road traffic, determining the trajectory is an essential task that requires the consideration of all relevant information pertaining to the situation at hand. In order to comprehensively evaluate the scene, we choose to employ trajectory prediction as an indirect approach.

\begin{figure}
    \centering
    \def\svgwidth{0.93\columnwidth}
    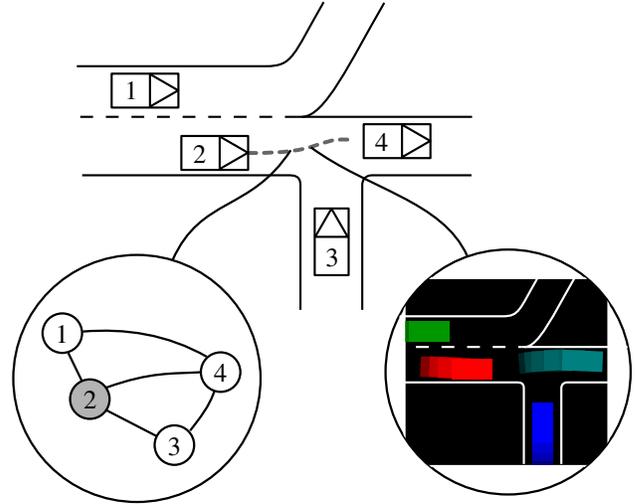
    \caption{Combination of a graph representation and a rasterized representation of the traffic scene for a trajectory prediction task}
    \label{fig:approach_overview}
\end{figure}

This study proposes a novel approach to identify significant traffic scenarios by employing two  separate machine learning methodologies based on two distinct representations of the environment (see \Cref{fig:approach_overview}). One path utilizes an image-based model that incorporates spatial information, particularly the road information, while the other employs a graph representation of the environment specifically designed to represent social contexts within a traffic scene.
The objective of this study is to quantify the level of interactivity in a traffic scenario based on the mutual influence of the two models. The proposed approach should make it possible to make a statement about the degree of interactivity in a given traffic scenario.

This paper is structured as follows: In \Cref{sec:sota} we provide a review of existing traffic scenario descriptions and related work in the context of motion prediction.
\Cref{sec:scene_description} describes a short overview over the used scene representation model and how certain attributes of the graph are derived. 
In \Cref{sec:architecture}, our implementation in regard to the used datasets is discussed. Moreover, two network architectures are proposed on how to extract information from a scene graph and an image. The results of the models are shown and evaluated in \Cref{sec:experiments}.
Finally, in \Cref{sec:conclusion} we conclude this contribution.

% ===================
%  State of the art
% ===================
\section{State of the Art}

\label{sec:sota}
Trajectory prediction, and especially criticality assessment of traffic scenes, is a topic that receives a significant amount of attention in research. In this section, we will introduce relevant representatives and approaches of trajectory prediction and then give a brief overview of traffic scene evaluation metrics.

In early attempts of driving trajectory prediction, manually designed features or heuristic methods for classifying specific driving manoeuvres were often used \cite{lefevre_survey_2014}. Later, Neural Networks, especially Recurrent Neural Networks (RNN) based on Gated Recurrent Units (GRU) or Long-Short-Term-Memory (LSTM) were designed \cite{deo_multi-modal_2018}. Current methods are largely based on supervised learning, but often complex energy functions serve as loss functions or multiple loss functions are involved whose influence must be determined manually. Two types of architectures can be distinguished here: On the one hand, networks that operate on synthetic images. Convolutional Neural Networks (CNN) are usually used to introduce image-encoded context information into the model. Subsequently, Recurrent Neural Networks (RNN) or transformers are usually applied
 \cite{krizhevsky_imagenet_2017, lee_desire_2017, tang_multiple_2019, liu_multimodal_2021, bansal_chauffeurnet_2018, chen_deep_2019}.
On the other hand graph-based models encode vehicles or lane segments as nodes. They utilize graph convolutions to propagate dynamic information over multiple time steps through their graphs.

Huang et al. \cite{huang_survey_2022} provide a comprehensive discussion of trajectory prediction for autonomous vehicles in their paper. Further notable works on this topic are mentioned below.

\subsection{Trajectory prediction with image-based networks}
In the area of image-based trajectory prediction, Bansal et al. \cite{bansal_chauffeurnet_2018} use a simple feature CNN to extract features from an abstract bird's eye view image representation. These are then further processed in an RNN that predicts an agent's movements. Parallel RNN steams offer different loss information using both vehicle and context information.

The authors of \cite{chen_deep_2019} take a similar but even simpler approach to design a vehicle controller. Their model is a pure feed-forward CNN, namely the VGG16-Net \cite{simonyan_very_2015}, which was originally developed for purposes of classification and detection in computer vision and adopted as such. %The model is trained solely on basis 
Zeng et al. \cite{zeng_dsdnet_2020} postulate a CNN-based model to represent probability distributions of traffic participant's movements from vehicle sensor data for safe vehicle movement planning.

\subsection{Trajectory prediction with graph-based networks}
Rico et al. \cite{rico_graph_2021} evaluate the suitability of Graph Neural Networks for vehicle trajectory prediction. Here, vehicles are represented as nodes connected to neighbouring vehicles by edges. The authors compare Graph Convolutional Networks and Graph Attention Networks and investigate different improvements. Among other aspects, they find that a graph with purposefully chosen edges performs superior to a fully meshed graph underlining the importance of graph topology.

Gilles et al. \cite{gilles_gohome_2021} form a graph of connected partial lanes, called lanelets. These are encoded by a recurrent network and their features are propagated through the graph using graph convolutions. Vehicle trajectories are analogously encoded and updated by the activations of the lanelet subnetwork. Using a self-attention mechanism, agent interactions are incorporated. Combining vehicle interactions and road context, they reach leading performance in the INTERPRET challenge benchmark \cite{noauthor_interpret_nodate}.

Pan et al. \cite{pan_lane-attention_2020} classify trajectory prediction approaches by their focus on dynamic interaction in contrast to static environment interactions and the use of Euclidean vs. non-Euclidean reference systems. In their model, they incorporate social interaction as well as environmental context using a graph structure where both vehicles and lanes are represented as nodes. Corresponding vehicle and lanes nodes are connected by edges. Temporal relations are encoded as edges as well. Information propagation here is led by spatial proximity of traffic participants as well as embedded historical information.

Yu et al. \cite{yu_spatio-temporal_2020} design an architecture for pedestrian movement prediction. Their method relies on the use of spatial and temporal transformers representing pedestrians as graph nodes using multi-head attention. Pedestrians move much more dynamically than vehicles. It can thus be concluded that social interactions are the dominant driver of behaviour here.
However, Syed and Morris \cite{syed_stgt_2021} successfully combine Spatio Temporal Graph Neural Networks, image-based feature extraction using CNN, and a transformer network for pedestrian trajectory prediction. They achieve significantly better results on the relevant comparison datasets than Yu et al. It is reasonable to assume that this is due to the incorporation of information about the static environment using graph processing.

The approach proposed by Gao et al. (Vector Net) \cite{gao_vectornet_2020} leverages an environment representation comprising both map and traffic participant features to facilitate the learning of ego trajectories. To this end, all relevant features are converted into vectors that have both a start and an end point. These vectors are then combined in a fully connected graph neural network (GNN) architecture, wherein the node features are defined by the start and end points of the vectors, as well as the timestamps of the trajectories, among other relevant factors.
This approach has been extended to target-based prediction \cite{zhao_tnt_2020} and anchor-free dense goal sets \cite{gu_densetnt_2021}.

% ===================
%  Methodology
% ===================

\section{Scene Representation}
\label{sec:scene_description}
In order to represent different aspects of the environment and to use them in the trajectory prediction, we exploit two different types of environmental representations. On the one hand, a graph representation to specifically depict social interactions between traffic participants and, on the other hand, a raster-based method for the geometric description of the environment is applied.

As a data basis for the investigation carried out in this work, object list-based motion datasets and the corresponding HD-maps are utilized. For the later evaluation, we use the version 1.2 of the INTERACTION dataset \cite{zhan_interaction_2019}.
\subsection{Graph Representation}
\begin{figure}[tbp]
  \centering
  \def\svgwidth{\columnwidth}
  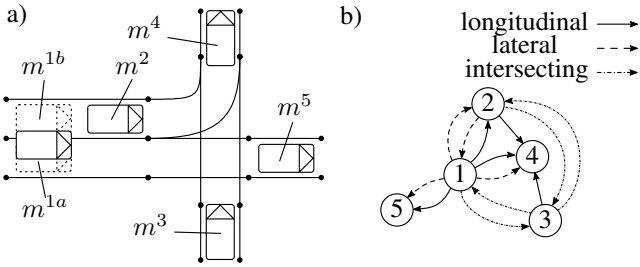
  \caption{Five vehicles projected to near lanes resulting in six projection identities $m$ (a). The resulting scene graph, where each traffic participant is represented by a node and the relations between its projection identities as edges (b) \cite{zipfl_towards_2022}.}
  \label{fig:scene_graph}
\end{figure}

The graph representation used is based on the Semantic Scene Graph (SSG) by Zipfl et al. \cite{zipfl_towards_2022}. A large part of the geometric information is discarded, and the focus is on the topological information. In \Cref{fig:scene_graph}, an exemplary traffic scene (a) and the resulting graph (b) are shown.
A traffic participant $i$ is represented as nodes $v^i$ in the graph. The relation $e^{ij}$ between two traffic participants ($i,j$) is defined based on the respective position of the traffic participants on the road (projection identity $m$), under consideration of the road topology.
This means that traffic participants, when driving one after the other on the same lane, get a longitudinal relation in the scene graph (see e.g. vehicle 2 and vehicle 4 at \Cref{fig:scene_graph}b) in addition to the type of the relation, the distance between the traffic participants along the lane is stored in the relation. Furthermore, traffic participants can have a lateral relation to other entities if they are on adjacent lanes or an intersection relation type if the entities drive on intersecting lanes.
In principle, the SSG maps the traffic scene into the Frenet space described by the road geometry.
For further information regarding the scene graph's implementation and the calculation of both the projection identities and the relations we refer to the paper \cite{zipfl_towards_2022}.

In the original implementation of the SSG, it is intended that there are parallel edges between two traffic participants that have multiple projections identities. In this work, parallel edges are merged into one main edge. State probabilities of the individual types are combined as probability value $[0,1]$ in the edge with the respective type.

The features in \Cref{tab:features} for any node $v^i$ and an edge $e^{ij}$ are thus used as input variables for our machine learning approach.\\[1ex]

\begin{table}[htb]
\caption{Description of the node and edge features utilized for training.}
\label{tab:features}
\centering
\begin{tabular}{@{}p{2.5cm}p{5.5cm}@{}}
\toprule
    Node Feature &\\
 \midrule
  velocity & Norm of the objects' velocity vector.\vspace{0.5ex}\\
  object class & One-hot encoding of the object type [car, truck, pedestrian, bike]\\
 \midrule
    Edge Feature &\\
 \midrule
 lat/lon/int certainty & Normalized certainty that the origin object has a longitudinal/lateral/intersecting relationship with the target object\vspace{0.5ex}\\
 path distance & Distance between the origin object and the target object in Frenet space if there is a lateral or longitudinal relationship (certainty $> 0$).\vspace{0.5ex}\\
 int path distance & Distance from the origin object to the intersection of its own lane and the lane of the target object, if the intersecting certainty $> 0$\vspace{0.5ex}\\
 centerline distance & Distance to the centerline of the lane of the origin projection identity with the highest lateral or longitudinal certainty\vspace{0.5ex}\\
 int centerline distance & Distance to the centerline of the lane of the origin projection identity with the highest intersecting certainty\\ 
 \bottomrule
\end{tabular}
\end{table}

\subsection{Image Representation}
The raster-based environmental representation's synthetic images are based on the recent works of \cite{bansal_chauffeurnet_2018} and \cite{chen_deep_2019}. 

Trajectory prediction is carried out for a specific ego vehicle, with a corresponding view of the traffic scene generated to center on the ego vehicle, which is consistently depicted in red with the front of the vehicle oriented towards the right of the image. Other traffic participants are illustrated as randomly coloured rectangles using only the other two colour channels. White lines denote lane edges and gray lines depict other lane markings. Additionally, virtual lane boundaries that are not visible in reality are represented by blue dashed lines, which signify possible turns at intersections, for instance.
Aside from the current position of each traffic participant, their movement history is also displayed using fading rectangles, where the degree of fading corresponds to the temporal distance from the present state. The motion history is discretized into ten states over a period of one second.

\begin{figure}[htbp]
  \centering
  \includegraphics[width=0.6\columnwidth]{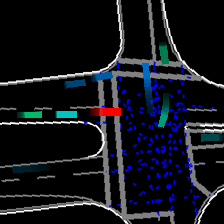}
  \caption{An example of the image representation of a traffic scene. The ego vehicle is depicted in red and the image section is chosen in such a manner that the vehicle is centered in the image and directed exactly to the right. Vehicle's previous locations are represented by fading colours.}
  \label{fig:image_example}
\end{figure}

The edge of each square birds-eye scene is 53\,m, which, from our point of view, contains all the information about the environment that is important for the ego-vehicle. The resolution of the input image is 224\,px$\times$224\,px which is given by the applied VGG network, so each pixel covers 0.25\,m~$\times$~0.25\,m in reality. 
The image is encoded in RGB format, with black pixels representing the background areas that contain no explicit information. An example of this data representation is given in \Cref{fig:image_example}.

% ===================
%  Methodology
% ===================
\section{Model Architecture and Training}
\label{sec:architecture}
\begin{figure*}[t]
  \centering
  \def\svgwidth{\linewidth}
  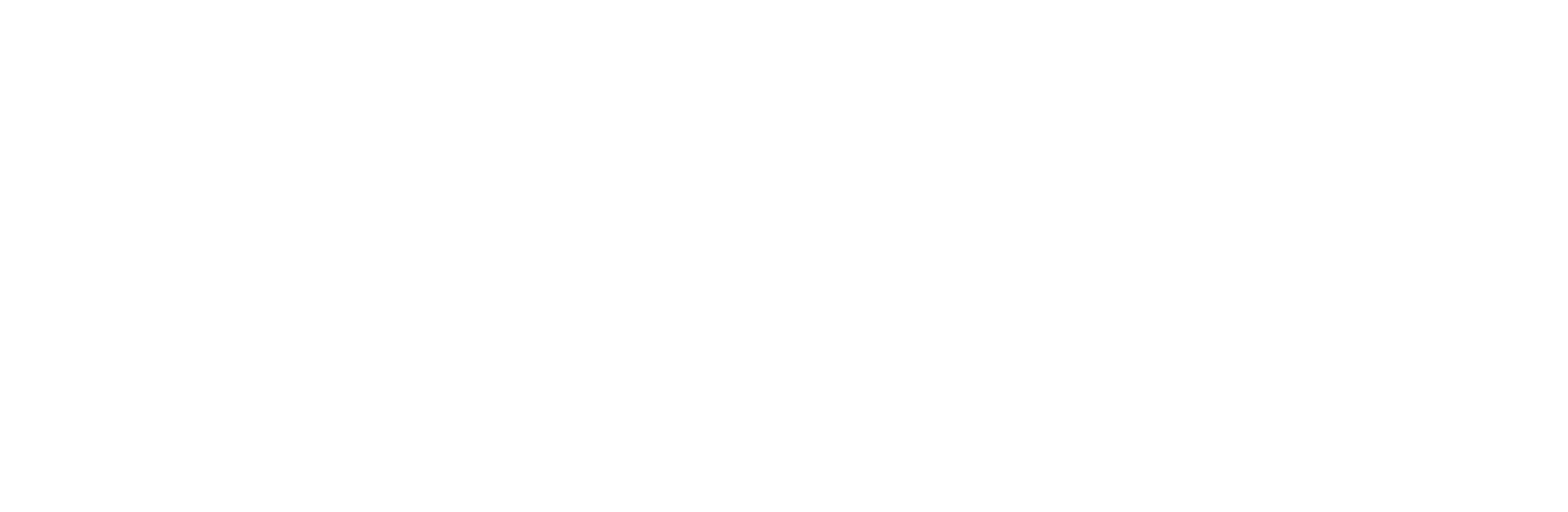 
  \caption{Trajectory prediction architecture for an ego-vehicle $i$, consisting of two parallel models for the computation of both graph ($G(t)$) and image information ($I^i$). Using a learned attention factor ($\alpha_I^i, \alpha_G^i$), the resulting embeddings ($\epsilon^i_G, \epsilon^i_I$) are combined.}
  \label{fig:hybrid_architecture}
\end{figure*}

Given a sequence of observed states (history) of length $T$, our goal is to predict the trajectory $\hat{\mathcal{T}^i}$ of an ego-vehicle $i$.
The observed states not only encapsulate the ego-vehicle's information but also that of its surroundings, which comprises the poses and velocities of all neighbouring traffic participants, and is further augmented and contextualized by the road information.

In \Cref{fig:hybrid_architecture} our hybrid architecture to predict the future trajectory $\hat{\mathcal{T}}^i$ of an ego-vehicle $i$ is depicted.

The architecture is composed of two parallel model branches, the graph pipeline and the image pipeline, each of which calculates an independent embedding, which are combined at the end. Relative influence of both branches is determined using an attention mechanism. The overall model is trained as a whole.

On the graph branch the first step is to compute, for each node $v^i$ in each graph $G(t)$ representing a social scene, an embedding that contains information about the social neighbourhood $\mathcal{N}^i$ of $v^i$.
This is achieved with the help of a Graph Convolutional Network (GCN) via message passing. 
We use an edge-conditioned convolution as described in \cite{simonovsky_dynamic_2017} and \cite{gilmer_neural_2017} where neighbouring node's features are mapped by a matrix that is determined by a learned function that uses corresponding edge features as inputs before message passing.
Specifically, we apply one graph convolutional layer meaning that vehicle behaviour estimation takes into account its direct neighbours in the scene graph. The node feature size is contained at its input size of 5 while the edge dependent transformation matrix is computed using a Multilayer Perceptron (MLP) with one hidden layer and LeakyReLU activation. As an aggregation function we use the mean function.
Resulting node features are further processed by two activated fully connected layers as it is suggested in \cite{rico_graph_2021}.

The result is a node embedding $\epsilon^i(t)$ for each traffic participant $i$ for each time step $t$ in the scene. 
For further computation of each traffic participant $i$ only the corresponding embedding $\epsilon^i_G$ is taken.

For each traffic participant the embeddings of the historic states are then combined into one predicted state using a two-layered GRU network with layer sizes equalling the input and output size.
This results in an intermediate representation of information about the vehicle's future trajectory as a 128-dimensional vector.
However, given the nature of the graph's input data, this representation is not sufficient to be transformed into an accurate future trajectory. This is due to the non-Euclidean nature of the processed edge features as well as the use of the norm of the velocity vector as a node feature. This ensures that only information that can be deducted by the behaviour of other traffic participants is contributed by this model branch.

The image data is processed through the VGG16 network \cite{simonyan_very_2015}. A linear layer is appended to produce a 128-dimensional image embedding $\epsilon_I^i$ for the corresponding vehicle $i$. 
To ensure equal weighting, both embeddings are normalized before being separately processed by an attention network using shared weights.
The attention network generates two scalar values, $\alpha_I$ and $\alpha_G$, indicating the importance of either $\epsilon_I^i$ or $\epsilon_G^i$ with respect to trajectory prediction.
Furthermore, $\alpha_I^i$ and $\alpha_G^i$ are bounded to a range of $[0.1, 0.9]$, while also ensuring that their sum is 1.
By applying these constraints, the individual embeddings are guaranteed to contribute at least 10\% to the final embedding. This is crucial for the training process, as it ensures that the slower training of either sub-model is not stalled prematurely.
The combined embedding $\epsilon^i_C$ is obtained by a linear combination of the attention factors and the respective model embeddings.
\begin{align}
    \label{eq:linear_combination}
    \epsilon_C^i = \alpha_G^i \epsilon_G^i + \alpha_I^i \epsilon_I^i
\end{align}
Following the embedding calculation, a two-layer, fully connected MLP with LeakyReLU activation function ($\theta_{traj}$) is utilized to generate the final velocity sequence. The velocity sequence corresponds to the trajectory of the ego-vehicle and is represented by discrete $\hat{\dot{x}}$ and $\hat{\dot{y}}$ components at time intervals ranging from 0 to $\hat{T}$:
\begin{align}
    \label{eq:trajectory}
    \hat{\mathcal{T}^i} = (\hat{\dot{x}}_{0}, \hat{\dot{y}}_{0}, \hat{\dot{x}}_{1}, \hat{\dot{y}}_{1}, ..., \hat{\dot{x}}_{\hat{T}}, \hat{\dot{y}}_{\hat{T}}).
\end{align}

\subsection{Training}
The dataset was partitioned into three distinct sets for training purposes. The first set, comprising 10\% of the data, was reserved as a holdout set for subsequent evaluation. The second set, accounting for 5\% of the data, was designated as the validation set. The remaining and largest set, encompassing 85\% of the data, was utilized for the actual training process. The model was trained using the loss function $\mathcal{L}$ specified in \Cref{eq:loss} and the Adam optimizer. The initial learning rate is set at $1\cdot10^{-4}$ and was subject to occasional reduction by one order of magnitude whenever the validation batch failed to progress. Using an Nvidia GTX 2080 Ti GPU a batch size of 48 is employed. To avoid RAM size limitations training data is stored in a compressed format and loaded from disk as needed. Each training iteration required approximately 60 epochs until no further improvement could be achieved.

\begin{align}
    \label{eq:loss}
    \mathcal{L}^i = \sum_j |\hat{\mathcal{T}}_j^{i}-\mathcal{T}_j^i|
\end{align}

% ============
%  Evaluation
% ============
\section{Evaluation and Experiments}
\label{sec:experiments}
The evaluation of our model was conducted through several methods. Firstly, given its function as a trajectory predictor.
Secondly, we assessed the performance of sub-strands of our model for various road types. In addition, a qualitative analysis was employed to explore how the attention parameters could be utilized to extract pertinent scenarios from a given dataset.

\subsection{Trajectory Prediction}
Each sample of the dataset consists of $T=1$\,s history and 3\,s future trajectory to be predicted. Each sample is discretised into 40 individual time steps.
To facilitate comparisons between our prediction model and other state-of-the-art models, we expanded the architecture to predict not just one trajectory but six different modes. This extension aligns with the INTERPRET Challenge \cite{noauthor_interpret_nodate} standard, enabling meaningful comparisons of the results .

Our models are evaluated using the Minimum Average Displacement Error (ADE), Minimum Final Displacement Error (FDE) and Miss Rate (MR) as proposed by the INTERPRET challenge benchmark. ADE describes the average Euclidean distance of proposed vehicle locations to the ground truth over the predicted timesteps in meters. The FDE is the displacement of the last predicted position and the Miss Rate the proportion of predictions that fall below certain velocity-dependent thresholds computed for longitudinal and lateral displacement separately. Generally, more than one trajectory prediction is generated where only the best one is chosen for error computation.

\Cref{tab:prediction_results} presents a comparison between the hybrid network, which combines the graph model and the image network (see \Cref{fig:hybrid_architecture}), and the image network alone, across various road types and locations. As expected, the hybrid architecture outperforms the image network, likely due to its increased parameter count and explicit social interaction.

In terms of road type, the incorporation of social interaction in prediction demonstrates the least relative impact for highway and merging scenarios, resulting in an average improvement of 71\%. Whereas, roundabouts exhibit an average improvement of approximately 91\%, while intersections fall between merging scenarios and roundabouts, with a 78\% average improvement. This observed pattern persists across all road types examined, except for a road geometry labelled as roundabout with the name EP (Improvement: 76.46\%). However, upon closer examination of this road geometry, it becomes evident that it represents a hybrid of an intersection and a small roundabout, thereby aligning with the aforementioned pattern described earlier.

\begin{table*}[htbp]
\centering
\caption{Prediction Results of both the hybrid model and the image model}
\label{tab:prediction_results}
\begin{tabular}{@{}llccccccc@{}}
\toprule
Name    & Type & ADE (hybrid) & FDE (hybrid) & MR (hybrid) & ADE (image) & FDE (image) & MR (image) & Improvement (FDE)\\
\midrule

ZS0 	&	Merging	&	0.19	&	0.60	&	0.1	&	0.38	&	0.88	&	0.24	&	68.79\,\%		\\
MT	&	Merging	&	0.29	&	0.90	&	0.18	&	0.52	&	1.24	&	0.3	&	72.45\,\%		\\
LN	&	Roundabout	&	0.49	&	1.49	&	0.49	&	0.68	&	1.73	&	0.46	&	86.38\,\%		\\
OF	&	Roundabout	&	0.76	&	2.44	&	0.54	&	0.85	&	2.29	&	0.49	&	106.68\,\%		\\
EP	&	Roundabout	&	0.41	&	1.33	&	0.39	&	0.64	&	1.66	&	0.47	&	79.88\,\%		\\
																		
SR	&	Roundabout	&	0.29	&	0.96	&	0.3	&	0.43	&	1.08	&	0.39	&	89.33\%		\\
EP0	&	Intersection	&	0.36	&	1.18	&	0.4	&	0.63	&	1.54	&	0.52	&	76.46\%		\\
EP1	&	Intersection	&	0.42	&	1.37	&	0.39	&	0.63	&	1.61	&	0.46	&	85.12\%		\\
																		
MA	&	Intersection	&	0.42	&	1.42	&	0.44	&	0.86	&	2.14	&	0.6	&	66.31\%		\\
\midrule
Total	&	All	&	0.40	&	1.30	&	0.36	&	0.62	&	1.57	&	0.44	&	82.57\%		\\
%\midrule
Multimodal & All & 0.20 & 0.51 & 0.06 \\

\bottomrule
\end{tabular}%
% }
\end{table*}

Although our model is capable of delivering multiple diverse trajectory predictions, we evaluate the benefit of our architecture on the basis of only one modality. This is to reduce complexity in our analysis as it is difficult to assess the plausibility of further predictions given that any additional proposal can only improve the common scores. However, for comparison with state-of-the-art models, in \cref{tab:prediction_results} we give results for our net predicting six different trajectories as it practised in the INTERPRET challenge benchmark. It can be seen that our model performs in the same range as state-of-the-art models in this setup. It achieves an ADE of $0.2$\,m, FDE of $0.51$\,m and a MR of $0.06$.

\subsection{Scenario Assessment}
\begin{figure}[t]
    \centering
    \begin{minipage}[t]{0.3\linewidth}
    \includegraphics[width=\linewidth]{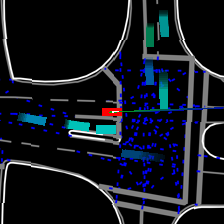}
    \end{minipage}
    \begin{minipage}[t]{0.3\linewidth}
    \includegraphics[width=\linewidth]{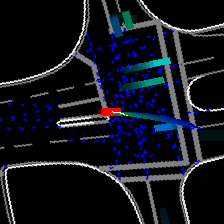}
    \end{minipage}
    \begin{minipage}[t]{0.3\linewidth}
    \includegraphics[width=\linewidth]{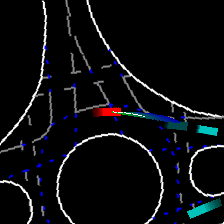}
    \end{minipage}
    \caption{Exemplary traffic situations and the corresponding trajectory prediction with a high graph attention value $\alpha_G$}
    \label{fig:high_graph_attention}
\end{figure}

In order to assess the traffic scenarios, we take into account the attention parameter. Specifically, as $\alpha_G$ approaches a value of 1, the graph component is weighted higher, resulting in a greater influence of the social component on the prediction of trajectories. To further illustrate this concept, we present \Cref{fig:high_graph_attention} which displays three exemplary instances of traffic scenes characterized by a particularly high value of $\alpha_G$. In addition to the ground truth trajectory depicted in white, the figure also presents an equal distribution of 50 sampled trajectories that are determined based on the degree to which the graph component (green: [$\alpha_G = 1, \alpha_I=0$]) and the image component (blue: [$\alpha_G = 0, \alpha_I=1$]) are included in the calculation.
These examples demonstrate the critical role played by the graph component in certain scenarios, such as braking to other traffic participants.

\begin{figure}[t]
    \centering
    \begin{minipage}[t]{0.3\linewidth}
    \includegraphics[width=\linewidth]{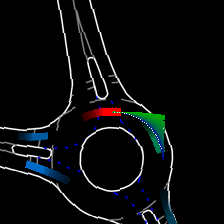}
    \end{minipage}
    \begin{minipage}[t]{0.3\linewidth}
    \includegraphics[width=\linewidth]{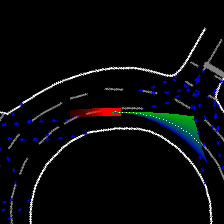}
    \end{minipage}
    \begin{minipage}[t]{0.3\linewidth}
    \includegraphics[width=\linewidth]{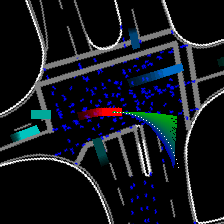}
    \end{minipage}
    \caption{Exemplary traffic situations and the corresponding trajectory prediction with a high image attention value $\alpha_I$}
    \label{fig:high_image_attention}
\end{figure}

The opposite effect can be observed when $\alpha_I$ is particularly high and $\alpha_G$ is correspondingly low. \Cref{fig:high_image_attention} depicts three traffic scenes featuring a high value of $\alpha_I$ for the ego-vehicle. It is noticeable that no other traffic participants are deemed relevant for trajectory calculation. Furthermore, the benefit of the image network is evident in this scenario, as the graph component lacks information about the road and consistently predicts a straight trajectory. In contrast, the image network attempts to follow the road.
This relationship between the curvature of the trajectory and $\alpha_G$ can be observed to a significant extent across the entire data set. 
Here $\alpha_G$ of a predicted trajectory is compared against the \emph{normalised curvature} of the trajectory's path.
In this case, the \emph{normalised curvature} is defined between $[0,1]$, where $0$ represents a straight path and $1$ represents a U-turn. The correlation coefficient in the examined dataset is $-0.452$ and shows an inverse correlation between $\alpha_G$ and the curvature of the trajectory. This means that large values of $\alpha_G$ result in trajectories with low curvature.

The application of $\alpha_G$ facilitated the identification of highly interactive situations through qualitative assessment. However, it is important to acknowledge that trajectories with high curvature may yield to a low $\alpha_G$ value, potentially leading to misclassification.

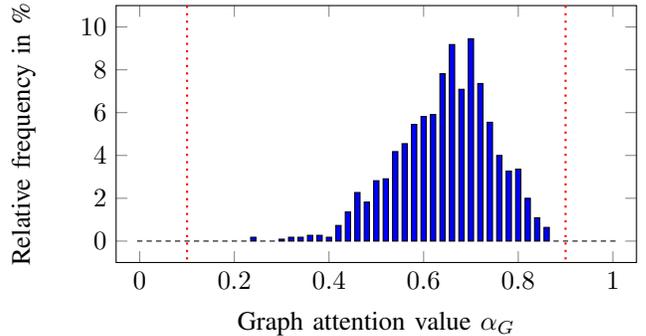
\begin{figure}[ht]
    \centering
    \begin{tikzpicture}
    \begin{axis}[
        xtick={0,0.2,...,1},
        xmax=1.05,
        xmin=-0.05,
        ytick={0,2,...,10},
        ymax=11,
        ymin=-1,
        bar width=2pt,
        xlabel={Graph attention value $\alpha_G$},
        ylabel={Relative frequency in \%},
        height=5cm,
        width=8.5cm,
      ]
        \addplot[ybar,fill=blue] coordinates {
(0, 0.00)
(0.02, 0.00)
(0.04, 0.00)
(0.06, 0.00)
(0.08, 0.00)
(0.1, 0.00)
(0.12, 0.00)
(0.14, 0.00)
(0.16, 0.00)
(0.18, 0.00)
(0.2, 0.00)
(0.22, 0.00)
(0.24, 0.18)
(0.26, 0.00)
(0.28, 0.00)
(0.3, 0.09)
(0.32, 0.18)
(0.34, 0.18)
(0.36, 0.27)
(0.38, 0.27)
(0.4, 0.18)
(0.42, 0.73)
(0.44, 1.36)
(0.46, 2.27)
(0.48, 1.82)
(0.5, 2.82)
(0.52, 2.91)
(0.54, 4.18)
(0.56, 4.55)
(0.58, 5.45)
(0.6, 5.82)
(0.62, 5.91)
(0.64, 7.82)
(0.66, 9.18)
(0.68, 7.09)
(0.7, 9.45)
(0.72, 7.36)
(0.74, 5.55)
(0.76, 4.00)
(0.78, 3.27)
(0.8, 3.36)
(0.82, 2.00)
(0.84, 1.09)
(0.86, 0.64)
(0.88, 0.00)
(0.9, 0.00)
(0.92, 0.00)
(0.94, 0.00)
(0.96, 0.00)
(0.98, 0.00)
(1, 0.00)
        };
        \addplot[thick, samples=50, smooth,domain=0:1,red,dotted] coordinates {(0.9,-5)(0.9,30)};
        \addplot[thick, samples=50, smooth,domain=0:1,red,dotted] coordinates {(0.1,-5)(0.1,30)};
    \end{axis}
\end{tikzpicture}
    \caption{The relative frequency distribution of graph attention values $\alpha_G$}
    \label{fig:histogram}
\end{figure}

\Cref{fig:histogram} shows the relative frequency distribution of the graph attention values $\alpha_G$ in the dataset. 
The definition range of the graph attention ($0.1 < \alpha_G < 0.9$), which is given by the training process, is indicated by the vertical red lines in the figure. In general, it is evident that, in the majority of cases, the attention is primarily focused on the graph model ($\alpha_G > 0.5$). Notably, the highest concentration of approximately 9.45\% of traffic scenarios falls within the range of 0.7-0.72. This distribution implies that social interaction among traffic participants or at least the information of their relationship, which is represented by a graph in our approach, plays a crucial role in most traffic scenarios and significantly influences the trajectory prediction.
Based on the outcome of our evaluation, we propose a recommendation emphasizing the significance of prioritizing the validation process for highly automated vehicles by placing special emphasis on scenarios characterized by a high $\alpha_G$ value.

% ============
%  Conclusion
% ============
\section{Conclusion}
\label{sec:conclusion}
This work introduces a novel approach for predicting trajectories in traffic scenes by employing a hybrid neural network architecture that incorporates rasterized image and graph input. Additionally, this methodology is used to determine the level of interactivity present in a given traffic scenario. It should be noted that our primary objective was not to attain the top-ranking performance metrics, but rather to showcase the effectiveness of using lightweight network architectures for scene classification purposes.

The interactivity of a traffic situation can be represented as a scalar, allowing for qualitative evaluation. However, a challenge remains as there is no definitive benchmark for interactivity and relevance in traffic scenarios. The findings of this study provide preliminary insights towards the eventual classification.

Future work needs to address the explicit link between the interactivity of a traffic scene and its relevance for validation.

% =================
%  Acknowledgement
% =================
\section{Acknowledgement}
The research leading to these results is funded by the German Federal Ministry for Economic Affairs and Climate Action" within the project “Verifikations- und Validierungsmethoden automatisierter Fahrzeuge im urbanen Umfeld". The project is a part of the PEGASUS family. The authors would like to thank the consortium for the successful cooperation.
\bibliographystyle{IEEEtran}
\bibliography{references_clean}

\end{document}